\documentclass[a4paper, 10pt, conference]{ieeeconf}      
\usepackage{FG2017}
\usepackage{graphicx}
\FGfinalcopy 

\IEEEoverridecommandlockouts                              
\title{\LARGE \bf
EAC-Net: A Region-based Deep Enhancing and Cropping Approach for Facial Action Unit Detection
}

\author{\parbox{16cm}{\centering
    {\large Wei Li$^1$   Farnaz Abtahi$^2$   Zhigang Zhu$^1$'$^2$   Lijun Yin$^3$ }\\
    {\normalsize
    $^1$ Grove School of Engineering, CUNY City College, NY, USA\\
    $^2$ Department of Computer Science, CUNY Graduate Center, NY, USA\\
    $^3$ Engineering and Applied Science, SUNY Binghamton University, NY, USA}}    
    \thanks{This work is supported by the National Science Foundation through Award EFRI -1137172, and VentureWell (formerly NCIIA) through Award 10087-12. The material is based upon the work supported in part by the National Science Foundation under grants CNS-1205664 and CNS-1629898.}
}

\begin{document}

\ifFGfinal
\thispagestyle{empty}
\pagestyle{empty}
\else
\author{Anonymous FG 2017 submission\\-- DO NOT DISTRIBUTE --\\}
\pagestyle{plain}
\fi
\maketitle

\begin{abstract}

In this paper, we propose a deep learning based approach for facial action unit detection by enhancing and cropping the regions of interest. The approach is implemented by adding two novel nets (layers): the enhancing layers and the cropping layers, to a pretrained CNN model. For the enhancing layers, we designed an attention map based on facial landmark features and applied it to a pretrained neural network to conduct enhanced learning (The E-Net). For the cropping layers, we crop facial regions around the detected landmarks and design convolutional layers to learn deeper features for each facial region (C-Net). We then fuse the E-Net and the C-Net to obtain our Enhancing and Cropping (EAC) Net, which can learn both feature enhancing and region cropping functions. Our approach shows significant improvement in performance compared to the state-of-the-art methods applied to BP4D and DISFA AU datasets. 

\end{abstract}

\section{INTRODUCTION}

\begin{figure}[thpb]
      \centering
      \includegraphics[scale=.5]{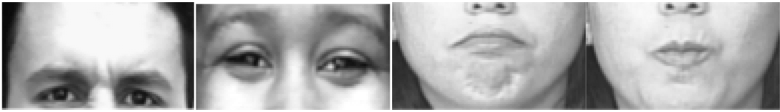}
      \caption{Action unit images for AU 4, 6, 7, 17}
      \label{fig1}
   \end{figure}

Facial Action Unit (AU) detection is an essential process in facial analysis. With a robust AU detector, facial expression and facial action problems can be solved more effectively. AU detection is the process to find some basic facial actions defined by FACS, the Facial Action Coding System \cite{p1}. Each AU represents a basic facial movement or expression change. Figure \ref{fig1} shows 4 basic AUs, namely eyebrows lower, cheek raiser, chin raiser and lip tighter.  The AUs are elements for more complicated facial actions. For instance, sadness might be the combination of AU1 (inner brow raiser), AU4 (brow lower), and AU15 (lip corner depressor). Most of current AU detection approaches either need the processed faces with frontal views or the texture features are artificially designed, making the features not well learned. To tackle these problems, we proposed the EAC (enhancing and cropping) Net to a convolutional neural network (CNN) to detect facial AUs automatically. { \color{black} We added the enhancing and cropping layers because even though a CNN has great capability in finding different patterns across images, it is not flexible enough to know which regions of the images need more attentions, and when comparing with peer images, the network is unable to shift the pixels to compare across corresponding regions. We enhance the regions of interest by assigning higher learning weights to corresponding areas during deep model training, and then crop corresponding areas to force the network to learn better representations by being trained on related regions.} We first build the enhancing net (E-Net), which is constructed by adding attention layers to a pretrained VGG net, one of the very effective CNNs. The E-Net yields a significant improvement in average F1 score and accuracy on BP4D dataset compared to the state of the art approaches. We then add cropping layers on top of the E-net and design the EAC Net. The cropping layers are implemented by cropping AU areas of interests from high-level convolutional feature maps. The EAC net yields up to 7.6\% increase in average F1 score and 19.2\% improvement in accuracy compared to the state of the art approaches applying to BP4D dataset. In addition to the improvement in metrics, our approach also has the following technical contributions: 
(1).	We propose an AU detection approach which is more robust to face position and orientation changes.
(2).	No facial preprocessing such as normalization is required to apply to the input images in our approach, which not only saves lots of preprocessing time, but also maintain the original facial expressions.
(3).	Although face landmarks are used in our EAC Net, they do not need to be very accurately located, i.e. the approach is robust to landmark detection errors.

 \begin{figure*}[thpb]
      \centering
      \includegraphics[scale=.65]{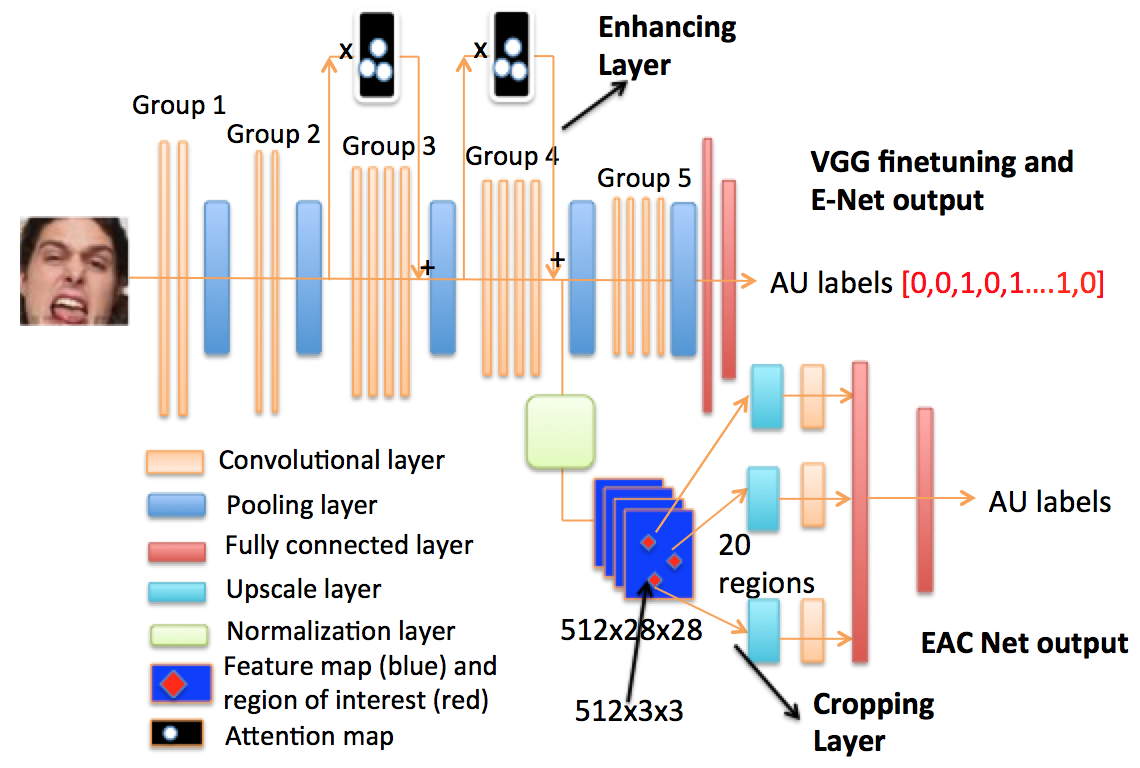}
      \caption{The EAC Net structure}
      \label{fig2}
   \end{figure*}
The idea of our approach is inspired by recent breakthroughs in deep learning research. Pretrained ImageNet models and transfer learning have found significant applications in many areas \cite{p2, p3}. The main reason is that the low-level convolutional layers extract similar features across different recognition tasks, which means the learned filters can be transferred. 
Yang, et al \cite{p4} proposed to use an attention layer for finding interesting areas to provide better answers in a visual question-answering task. In salient object detection \cite{p5}, a salient map is used to describe the important sub-areas of an image. This salient map (or attention map) is a 2D matrix with its elements ranging from 0 to 1, depicting the importance of corresponding pixels on the entire image. We believe that by applying an attention map to an existing network for facial computing, richer and more robust information can be obtained.  
The SPP net \cite{p6} proposed a ROI pooling idea, which can turn different interest areas into fixed length features. Faster RCNN \cite{p7} generated region proposals for objects to be detected and used the proposals to determine ``objectiveness". These findings made us believe that it might be possible to apply similar operation to facial interest regions and learn individual filters for specific regions. 
Thus, we designed our EAC net with three elements: a pretrained model, enhancing layers and cropping layers.

Figure 2 shows the structure of our EAC Net. The EAC Net includes 3 main parts. The first part is a fine-tuned pretrained VGG 19-layer network \cite{p8}. The low level convolutional layers (group 1 and 2) of the pretrained network and their parameters are kept for low-level visual feature extraction. The parameters of the higher-level convolutional layers (group 3 and 4) are updated for AU detection task. The use of VGG pretrained net is to guarantee that the network has a deep understanding of the input images. The second part of the EAC Net consists of the enhancing layers that are on top of group 3 and 4 convolutional layers. The purpose of adding these layers is to give more attention to individual AU areas of interest. The features extracted after the enhancing layers are supposed to contain more valuable information to detect AUs. The third part of the framework includes the cropping layers. Sub-features are cropped from ten selected interest areas of the feature map, followed by upscale layers and convolutional layers in each sub-area for further learning. The purpose of the cropping layers is to assure that only corresponding regions are compared by the network. Adding cropping layers as the higher convolutional layers would also help the region obtain deeper contextual information.

This paper is organized as follows. Related work is reviewed in Section 2. Section 3 introduces our approach and describes the details of EAC net. The experimental results are summarized in Section 4. Finally the conclusions are provided in Section 5.

\section{related work}

AU detection has been studied for decades and several approaches have been proposed. Facial key points play an important role in AU detection. Many conventional approaches \cite{p10, p11, p14,  p16, p17, p18, p24, p30} were designed by employing texture features near the facial key points. Valstar, et al. \cite{p9} analyzed Gabor wavelet features near 20 facial landmark points. The features were then selected and classified by Adaboost and SVM classifiers. Since landmark-based geometry changing is robust in many AU detection methods, Fabian, et al. \cite{p12} proposed an approach for fusing the geometry and local texture information. The geometry information is obtained by measuring the normalized facial landmark distances and the angles of the Delaunay mask formed by the landmark points. On the other hand, the texture features were obtained by applying multiple orientation Gabor filters to the original images. Zhao, et al. \cite{p13} proposed the Joint Patch and Multi-label Learning (JPML) for AU detection. Similarly, landmark-based regions were selected and SIFT features were used to represent the local patch. Overall, the conventional approaches focused on designing more representative features and finetuning more robust classifiers.  
In addition to facial AU detection, some researches also have focused on other related problems. Song, et al. \cite{p23} investigated the sparsity and co-occurrence of action units. Wu \cite{p19} exploited the joint of action unit detection and facial landmark localization and showed that the constraints can improve both AU and landmark detection. Girard, et al. \cite{p20} analyzed the affect of different sizes of training datasets on appearance and shape-based AU detection. Gehrig, et al. \cite{p21} tried to estimate the action unit intensity by employing linear partial least squares to regress intensity in AU related regions. 	

Over the last few years, we have witnessed that CNNs boost the performance in many computer vision tasks. Compared to most conventional artificially designed features, CNNs can learn and reveal deeper information from the training images which in turn contribute to better performance. Zhao, et al. \cite{p22} proposed a region CNN-based approach for AU detection. Instead of directly applying a regular CNN to the entire input image, the network divides the input image into 8x8 blocks and then trains over these regional areas independently. All sub-regions are then merged back into one net, followed by regular convolutional and fully connected layers. The proposed approach outperforms both the regular and Fully Convolutional Net (FCN).
Our proposed approach is also CNN-based and we share similar ideas with Zhao et al \cite{p22} in training sub-regions independently. The distinction of our approach is that instead of directly dividing the image into blocks, we use a ``smarter" way to find the important areas and crop the regions of interest. In their approach, the facial landmarks play an important role for normalizing the faces. Errors will accumulate in both landmark detection and face normalization processes, and facial normalization may neutralize expressions. However, in our approach, the network directly works on the interest areas, and even though landmarks are also used for building the attention map, our approach has large tolerance for landmark shifting, since we use a relatively large local region to cover the AU target areas. This will reduce errors from misalignment along images and will focus more on interest regions. 

\section{The EAC Net}

The EAC Net is composed of three parts: the VGG finetuning network, E-Net and C-Net. For comparison purposes, we implement three networks using VGG net as their base: F-VGG: the fine-tuning network of VGG; E-Net, the Enhancing Net based on the finetuned VGG; and finally EAC: the integration of Enhancing and Cropping Nets based on the trained E-Net model.

\subsection{F-VGG: VGG finetuning model}
Fintuning pretrained models for image classification are proved to be efficient in many areas \cite{p2, p3}. To make sure we can have a deep understanding of the images for AU detection, we employed the VGG 19-layer model. The VGG model follows a conventional CNN structure, comprising 16 convolutional layers and 3 fully connected layers. It also includes 5 pooling layers downsampling input images from 224x224 to eventually 7x7. In our implementation, we divide the convolutional layers into 5 groups as shown in Figure 2, which are separated by the 5 pooling layers. There are 2, 2, 4, 4, 4 convolutional layers in groups 1 through 5, respectively. The F-VGG structure is straightforward and has been modified in many related tasks such as object detection or emotion recognition.  Before designing our special purpos networks (E-Net and EAC-Net) for AU detection, we first modify and finetune the VGG net as our baseline approach. We keep the parameters of the first 3 groups of convolutional layers unchanged and update the rest of the layers during training. In order to match with the AU detection task, the number of nodes for the last 2 fully connected layers are changed to 2048 (by reducing parameters) and 12 (by matching the 12 AU targets). Dropout is also applied on both new layers to prevent overfitting during training.   

\subsection{E-Net: The Enhancing Net}
We mentioned that the idea for the enhancing net (E-Net) is inspired by the attention map and the salient map in the literature. Figure 2 demonstrates how the E-Net works by using enhancing layers. The feature map output from group 2 is multiplied by the designed attention map -- the first enhancing layer (details will be discussed below), in parallel with the convolutional layers in group 3. The two feature maps -- one from the enhancing layer and the other from the group 3 convolutional layers --  are then fused by element-wise summation.  Same operation is performed jointly by the second enhancing layer with the convolutional layers in group 4.  The reason why we designed the enhancing layer is that not all the areas of a facial image are equally important for individual AU detection. We can see that different AUs focus on corresponding sub-areas of the face. For example, in Figure \ref{fig3}, the eyebrow raiser AU is close to the texture in the area near the middle of the eyebrows and the lip corner AUs are determined by the texture information around lip corners. These areas are more important than the nose or most of the other parts of the face. For this reason, we build the attention map for AUs based on key facial landmarks, as shown in Figure \ref{fig3}. 

 
{\color{black} We have noticed that many previous works provide robust facial landmark positions \cite{p25}. Furthermore, our approach does not require the localization of landmarks to be very accurate in pixels since we are trying to find the areas for AUs.  We will work on 12 AUs as listed in Table \ref{tab1}, since these are the ones labeled in the datasets we use.  
After obtaining the facial landmarks as shown in Figure \ref{fig3}, we can define the centers for AUs and then build a bounding box around the center. Observing the AU figure, we manually define the center of AUs (the green spots) based on the muscles of a human face. Note that many AU centers are not directly on the same spots of the detected landmarks. 
We define a scaled distance as a reference for facial pixel shifting by calculating the distance of the corners of the two eyes, d, as shown in Figure \ref{fig3}. Then the centers for 12 listed AUs in Table \ref{tab1} are illustrated in Figure \ref{fig3} (the green spots). Since most AUs are symmetric on a human face, we define one pair of points for each AU. We should note that some AUs share the same centers, such as the lip corner puller and the lip corner depressor. So finally we defined 20 AU centers on the face for the 12 AUs listed in Table \ref{tab1}. The rules for defining the centers of the 12 AUs are also illustrated in Table \ref{tab1}. After obtaining the AU centers, we can build the attention map based on center positions. }
\begin{table}
\caption{Rules for defining AU centers}
\label{tab1}
\begin{center}
\begin{tabular}{|c|c|c|}
\hline
AU index & Au Name & AU Center\\
\hline
1 & Inner Brow Raiser & 1/2 scale above inner brow \\
2 & Outer Brow Raiser & 1/3 scale above outer brow \\
4 & Brow Lowerer & 1/3 scale below brow center\\
6 & Cheek Raiser & 1 scale below eye bottom\\
7 & Lid Tightener & Eye center\\
10 & Upper Lip Raiser & Upper lip center\\
12 & Lip Corner Puller & Lip corner \\
14 & Dimpler & Lip corner\\
15 & Lip Corner Depressor & Lip corner\\
17 & Chin Raiser & 1/2 scale below lip\\
23 & Lip Tightener & Lip center\\
24 & Lip Pressor & Lip center\\
\hline
\end{tabular}
\end{center}
\end{table}
Figure \ref{fig3} shows how the attention map is generated. Given an image like the one shown in Figure \ref{fig3} (left), we first obtain the landmarks for the key points on the face, which are shown with blue points. Having the facial key points, we can obtain the AU centers by shifting a distance or directly using existing facial landmarks. The AU centers are illustrated with green points in Figure \ref{fig3}. The AU centers are in pairs due to the symmetry of human face, with each AU center corresponding to one or more AUs. To make the shifting distance more adaptable to all face images, we define a measurement reference for the shifting distance. Inner corner distance is used as the scaled-distance, as shown in Figure 3. This scaled-distance is used in Table \ref{tab1} to help locate the AU centers. We first resize the images to 100x100 to make sure the same scales are shared among all images. Then, for each AU center, we define the nearby 5 pixels belonging to the same area, therefore the size of each AU area is 11x11. Higher weight is assigned to the closer points to the AU center. The relationship follows the equation: 

 \begin{figure}[thpb]
      \centering
      \includegraphics[scale=.4]{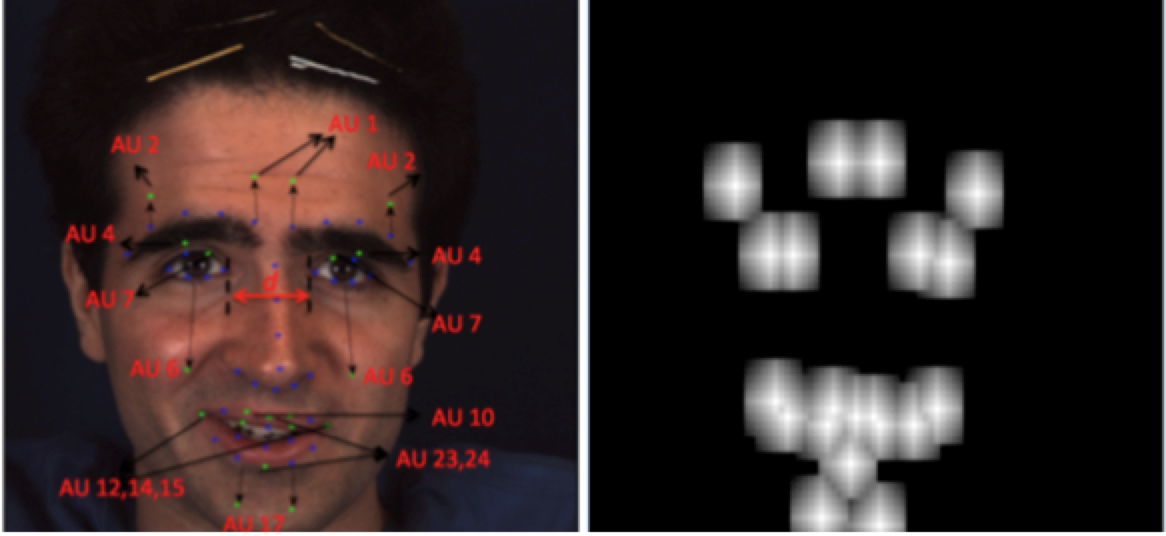}
      \caption{Attention map generation}
      \label{fig3}
   \end{figure}

\begin{equation}
w=1-0.095\cdot d_{m}
\end{equation}
where $d_{m}$ is the Manhattan distance to the AU center. 
 
An attention map obtained for a face image is demonstrated in Figure \ref{fig3}. The areas in the attention map with higher values correspond to the AU active area in the face image and can enhance deep learning at these areas in our enhancing layer. 
We can then apply the obtained attention map to the feature maps. Figure \ref{fig4} shows the E-Net structure. For group 1 and group 2, we keep the layers unchanged for detecting low-level features. For group 5 (size 14x14), the feature maps are too small to use any attention map, so eventually we apply the attention map only to group 3 and group 4, thus we add two enhancing layers. Adding attention layers directly to the feature maps by replacing the original ones will lose all the contextual information. So, we add the attention maps to the feature maps in parallel with the convolution operations, in group 3 and group 4, respectively, as shown in Figure \ref{fig4}. Element-wise summation is then conducted to obtain enhanced feature maps. We call our enhancing net based on the fine-tuned VGG model the E-Net. The E-Net structure is also similar to Residual Net [26], but is designed for generating enhanced features by applying an attention map.  The other structure in the E-Net is the VGG finetuning network. After training this model, we observed that the E-Net can lead to 5\% increase in average F1 score and 19\% increase in average accuracy on the BP4D AU dataset \cite{p27}. The detailed experimental results will be reported in the next few sections.

\subsection{EAC-Net: The Integrated Model with Enhancing and Cropping}
The E-Net can generate the features with more emphasis on the AU related regions, but it doesn't change the fact that the same AU areas are not normalized across images. To make sure the same regions are aligned across images, we use the approach in \cite{p22} which preprocesses all faces by normalization. Face normalization is a common preprocessing approach for face recognition. The disadvantage though is that during face normalization, some context information might be lost, and since face normalization is trying to align the faces to neutral expression, the facial expression might be weakened. The goal of our cropping net (C-Net) is to obtain each individual AU related area without changing the texturural information of the original images. Then, for each AU sub-area, independent convolutional layers are applied to learn more features.

Figure \ref{fig5} shows the structure of the C-Net. Cropping layers are added to the end of group 4, right after the enhancing feature maps are obtained. The output size for the feature map from group 5 is 512x28x28. With the same ratio as an AU area of 11x11 pixels in the attention map versus the face image of 100x100 pixels, the cropped areas should have the size of 3x3. For each of the 20 AU centers, we obtain a feature sub-map with size 512x3x3; in total we have 20 such feature sub-maps for the 20 AU centers. When adding convolutional layers after the cropped feature map, the new obtained feature map size will be 512x1x1. We feel that this is less representative for the AUs. So, before adding new convolutional layers, we apply an upscaling layer to the feature maps, upscaling the feature maps to 512x6x6. Actually our experiments show that this upscaling layer by itself leads to approximately 1\% increase in AUss average F1 score. 

To make the C-Net converge more quickly, we build the C-Net on top of the pretrained E-net, thus leading to the Enhancing and Cropping Net (EAC-Net). So the features obtained from group 4 have already been pretrained for AU detection. During implementation, we found that feature values obtained from the last convolutional layer of group 4 are very large and make the C-Net unable to converge. So a local response normalization layer is added before C-Net convolutional layers. The local response layer normalization algorithm follows equation \ref{att}. In our experiments, k=2, $\alpha$=0.002 and $\beta$=0.75. $x_{i}$ is a pixel among the 5 neighboring pixels. 
All the individual convolutional layers are followed by fully connected layers with a fixed size of 150. We then concatenate the fully connected layers. The rest is similar to F-VGG and E-Net.  
\begin{equation}
\label{att}
x_{i}=\frac{x_{i}}{(k+\alpha \Sigma _{i}x_{i}^{2})^{\beta} } 
\end{equation}

AU detection is different from regular classifications in the sense that instead of classifying images into one object category, multiple AUs can co-occur simultaneously. Thus this is a multi-label binary classification problem. Cross entropy as in \cite{p22} is used to measure the loss for this kind of problem.  In our loss function equation \ref{loss}, we added offsets to prevent the number from becoming too large.
\begin{equation}
\label{loss}
Loss=-\Sigma(l\cdot\log(\frac{p+0.5}{1.05})+(1-l)\cdot\log(\frac{1.05-p}{1.05}))
\end{equation}\

\section{Experiment}

\subsection{Datasets  and Evaluation Methods}

The most popular datasets for AU detection are CK+ \cite{p28}, BP4D\cite{p27} and DISFA\cite{p29}. AU datasets are harder to obtain compared to other tasks such as image classification. This is because there are multiple AUs in one face which requires much more manual labeling work. Here we give a brief review of the AU datasets referred and compared in this paper.

DISFA: 26 people are involved in the DISFA dataset. The subjects are asked to watch videos while spontaneous facial expression are obtained. The AUs are labeled with intensities from 0 to 5. We can obtain more than 100,000 AU-labeled images from the video, but there are much more inactive images than the active ones. The diversity of people also makes it hard to train a robust model. 

BP4D:  There are 23 female and 18 male young adults involved in the BP4D dataset. Both 2D and 3D videos are captured while the subjects show different facial expressions. Each subject participates in 8 sessions of experiments, so there are 328 videos captured in total. AUs are labeled by watching the videos, and the valid AU frames in each video varies from several hundreds to thousands. There are around 140,000 images with AU labels that we can use. 

To train a deep learning model, we need larger numbers of image samples, plus the diversity of the samples is also important. Similar to Zhao's experiment settings \cite{p22}, we choose BP4D to train our model. We first split the dataset to 3 folds based on subjects. Each time two folds are used for training and the third fold for testing. For the DISFA dataset, all samples are used for testing.
The balance of data is very important in training deep learning models. For our task of multi-label learning, this is even more challenging since several AUs are not independent of each other. The original occurrence rate for the 12 selected AUs is shown in the first row of Table \ref{tab2}. We can clearly see that the AUs are divided into 2 groups. AUs 6, 7, 10, 12, 14 and 17 are more representative than the minor AUs 1, 2, 4, 9, 11 and 12. If we just pick all the less representative AUs, the occurrence rate is shown in table \ref{tab2}, second row. 
We can see that even with only the less representative AU samples, the occurrence rate is imbalanced. We still need to keep the other samples to maintain the data diversity. Thus, we finally decided to try to keep the balance of training data by changing the selection rate during training. For all the training samples, we used to equally randomly pick a fixed number of images. To compensate for the less occurred AUs, we manually increase their rate during random picking operation by 4 to 7 times. Then the occurred rate is shown in Table \ref{tab2}, third row.  

\begin{table*}
\caption{BP4D samples balancing for AU occurrence}
\label{tab2}
\begin{center}
\begin{tabular}{|c|c|c|c|c|c|c|c|c|c|c|c|c|}
\hline
AU & 1 & 2 & 4 & 6 & 7 & 10 & 12 & 14 & 15 & 17 & 23 & 24\\
\hline
Original & 0.24 & 0.18 & 0.23 & 0.44 & 0.52 & 0.58 & 0.57 & 0.43 & 0.15 & 0.36 & 0.19 & 0.16\\
Minor AU occurred & 0.56 & 0.43 & 0.40 & 0.47 & 0.57 & 0.64 & 0.59 & 0.56 & 0.35 & 0.58 & 0.46 & 0.39\\
After balancing & 0.39 & 0.32 & 0.33 & 0.45 & 0.54 & 0.60 & 0.56 & 0.49 & 0.30 & 0.50 & 0.33 & 0.30\\
\hline
\end{tabular}
\end{center}
\end{table*}


{\color {black} Both the F1 score and the average accuracy are used to measure the performance of AU detection. In a binary classification scenario, especially when samples are not balanced, F1 score can better describe the performance of an algorithm \cite{p9,p10}. F1 score includes two components: precision (p) and recall (r). The precision is also called the positive predictive value and is the fraction of true positive predictions to all positive predictions. The recall is also called sensitivity and is the fraction of true positive predictions to all ground-truth positive samples.  Knowing p and r, we can obtain F1 using the following Eq. \ref{f1eq}}:

\begin{equation}
\label{f1eq}
F1=\frac{2p\cdot r}{p+r}
\end{equation}

\subsection{Implementation Details and Results}
In our neural network structures, we employed the basic VGG structure as the base, so all of the input images to the networks need to be resized to 224x224 to become compatible with the structures. In order to create the attention map for E-Net and interest region cropping parameters for the cropping layers, we use the facial landmark data provided by the dataset. We use a learning rate of 0.0001 in our training and a momentum of 0.9. First, VGG fine-tuned net is trained as the baseline for our proposed E-Net. EAC Net training is based on pretrained E-Net with new C-Net layers.  
We trained 3 models using the BP4D dataset: finetuned VGG (FVGG), E-Net, and EAC Net. The accuracy and F1 score for all 12 selected AUs and the average accuracy (i.e. the F1 score) are listed in Tables \ref{tab3} and \ref{tab4}. We also list the state of the art results from DRML \cite{p22} using deep learning and traditional approach JPML \cite{p13} in same settings for comparison.

\begin{table}
\caption{F1 score on BP4D dataset}
\label{tab3}
\begin{center}
\begin{tabular}{|c|c|c|c|c|c|c|}
\hline
AU& LSVM& JPML\cite{p13}& DRML\cite{p22}& FVGG& E-Net& EAC\\
\hline
1& 23.2& 32.6& 36.4& 27.8 &37.6& {\bf 39.0}\\
2& 22.8& 25.6& {\bf 41.8}& 27.6& 32.1& 35.2\\
4& 23.1& 37.4& 43.0& 18.3& 44.2& {\bf 48.6}\\
6& 27.2& 42.3& 55.0& 69.7& 75.6& {\bf 76.1}\\
7& 47.1& 50.5& 67.0& 69.1& {\bf 74.5}& 72.9\\
10&77.2&72.2&66.3&78.1&80.8&{\bf 81.9}\\
12&63.7&74.1&65.8&63.2&85.1&{\bf 86.2}\\
14&64.3&65.7&54.1&36.4&56.8&{\bf 58.8}\\
15&18.4&{\bf 38.1}&33.2&26.1&31.6&37.5\\
17&33.0&40.0&48.0&50.7&55.6&{\bf 59.1}\\
23&19.4&30.4&31.7&22.8&21.9&{\bf 35.9}\\
24&20.7&42.3&30.0&35.9&29.1&{\bf 35.8}\\
Avg&35.3&45.9&48.3&43.8&52.1&{\bf55.9}\\
\hline
\end{tabular}
\end{center}
\end{table}

\begin{table}
\caption{Accuracy on BP4D dataset}
\label{tab4}
\begin{center}
\begin{tabular}{|c|c|c|c|c|c|c|}
\hline
AU& LSVM& JPML\cite{p13}& DRML\cite{p22}& FVGG& E-Net& EAC\\
\hline
1&20.7&40.7&55.7&27.2&{\bf 71.1}&68.9\\
2&17.7&42.1&54.5&56.0&72.9&{\bf 73.9}\\
4&22.9&46.2&58.8&80.5&77.4&{\bf 78.1}\\
6&20.3&40.0&56.6&72.3&76.9&{\bf 78.5}\\
7&44.8&50.0&61.0&64.1&{\bf 70.7}&69.0\\
10&73.4&	75.2&53.6&72.4&75.7&{\bf 77.6}\\
12&55.3&60.5&60.8&69.1&82.8&{\bf 84.6}\\
14&46.8&53.6&57.0&52.8&56.7&{\bf 60.6}\\
15&18.3&50.1&56.2&67.4&77.6&{\bf 78.1}\\
17&36.4&42.5&50.0&61.2&69.3&{\bf 70.6}\\
23&19.2&51.9&53.9&72.2&80.2&{\bf 81.0}\\
24&11.7&53.2&53.9&77.0&82.3&{\bf 82.4}\\
Avg&32.2&50.5&56.0&64.4&74.5&{\bf 75.2}\\

\hline
\end{tabular}
\end{center}
\end{table}

As shown in Tables \ref{tab3} and \ref{tab4}, for the BP4D dataset, compared to the state of the art approaches, the VGG fine-tuned model (FVGG) has a higher average accuracy, but the average F1 score does not outperform the state of the art. Note that in our proposed approach, we do not perform any preprocessing to the input images. For the more representative AUs, the network is able to correctly predict the AU labels even without being told the position of the AUs. We believe that this is due to the depth of the VGG model, which can learn very deep features, and the pooling layers, which make the AU detection robust to position shifts. For the less representative AUs, the texture is less discriminative for instance around the eyebrow area. Also the occurrence rate is still smaller than the other AUs, making the training more challenging. 

\begin{figure}[thpb]
      \centering
      \includegraphics[scale=.4]{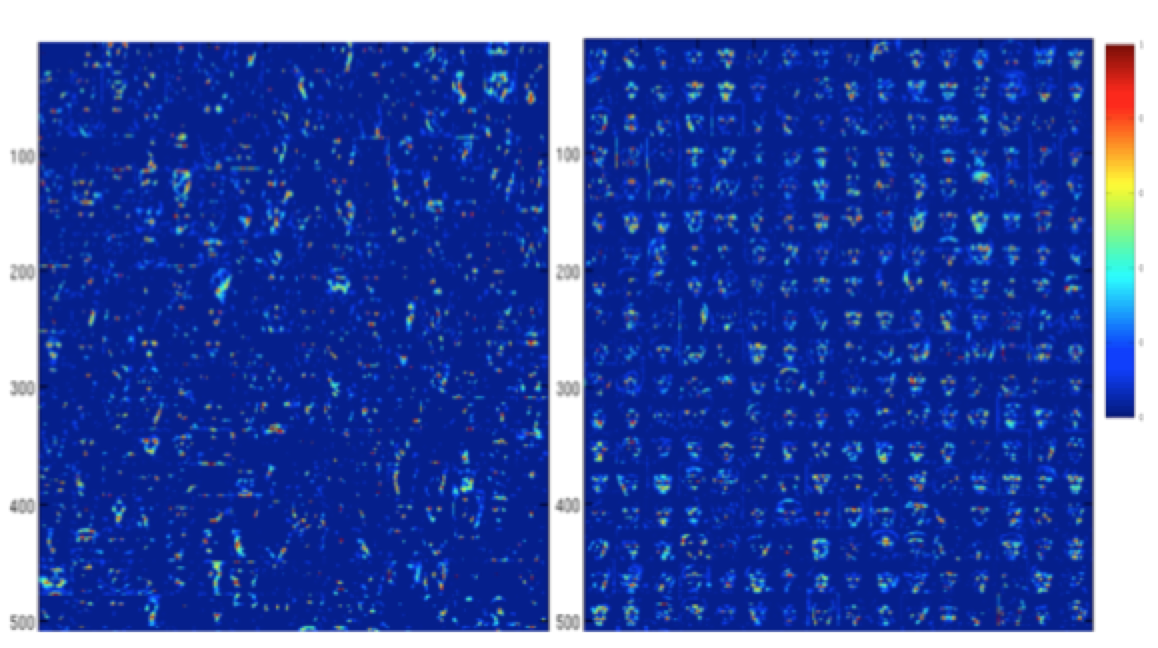}
      \caption{Visualization of selected feature maps from FVGG and E-Net.}
      \label{fig6}
   \end{figure}

The E-Net results show better average accuracy and F1 scores than the state of the art approaches and the VGG finetuning net.  On average, the improvement using E-Net over FVGG in F-score are 8.3\% and 10.1\%, respectively. The result can prove the effectiveness of our proposed enhancing layers. To explore more details of the E-Net, we extract the feature maps from multiple layers in the E-Net and VGG finetuning Net. The feature maps are illustrated in Figure \ref{fig6}.

We can see the attention map we designed made a big difference in the output feature maps. We plot the last feature maps of group 4 convolutional layers from the two structures, FVGG and E-Net. The feature map is 512x28x28. We mapped the feature map into an 896x448 image (height: 32x28 and width: 16x28). In the VGG feature map, the hot areas or the attention areas do not have a meaningful focus. In some areas, even the edges are highlighted as valuable features. The neural network has no idea which region to look into. While in the E-Net feature map, we can clearly see the network is concentrating on area on the face, mainly the regions enhanced by the attention map. This can make the E-Net extract more valuable features for AU detection.  

Finally, our EAC Net achieves the best performance in AU detection. Compared to the state of the art approaches, The EAC Net shows improvement in nearly all the AU detections results, except for AU2. We can see that the F1 score and accuracy all have a small improvement from E-Net to EAC Net. This means the pretrained E-Net performs well and the new structure of EAC net is able to extract more detailed information for AU detection. The improvement of the F-score and the average accuracy are 3.5\% and 0.7\% respectively.

\subsection{Discussions}

As we have seen in the experimental results on the BP4D dataset, the cropping layer of the C-Net only slightly increases the average accuracy, even though the F1 score, which is a more appropriate indicator of the performance of the algorithm, increases by 3.5\% over the E-Net. We know that the major role of the cropping layers is for alignment, but the faces of the BP4D data are mostly close to frontal view. We also have discussed that normalization is a typical processing step in AU detection in the literature. So we would like to answer two questions: (1) Will a preprocessing aligning step using normalization help or hurt the performance? (2) Will our C-Net increase the average accuracy of AU detection and the F1 score more significantly if the face images include more non-frontal views?
  
(1) Do aligned faces help or hurt?

We claim in our approach that the E-Net/EAC will automatically focus on the interesting regions and face alignment is not required. Plus, face alignment needs more computing power in real time applications. We conduct an experiment to compare the performance of the E-Net model on AU detection with aligned and non-aligned faces on random selected 2000 images from BP4D testing sets. In this experiment, both time and accuracy are considered.

For all the facial images, we first extract all the facial landmarks. For the non-alignment approach, E-Net is applied to directly find the AU prediction. For facial alignment approach, we run 2D alignment to register six 2D key points (eyes inner/outer corners, mouth corners) to a standard template. We then apply E-Net to the aligned face. The comparison is shown in Table \ref{tab5}. We randomly pick 3000 test samples from the test dataset. The data is tested in both aligned and non-aligned modes. The average time needed for landmark locating and normalization for each face image is also recorded. Average F1 score and accuracy (ACC) for the 12 AUs are used for comparison. From Table \ref{tab5} we observe that with the E-Net model, adding facial alignment to the images slightly hurts the performance. Even though this is not significant, the process of applying facial alignment is much more time-consuming, from only 30 ms to 2500 ms. For real-time implementation of the AU detection, the E-Net or EAC Net approaches can work well without additional preprocessing. Thus, we can make the conclusion that aligning faces would not contribute to the AU detection performance, but would cost much more time. So, our proposed EAC or E-Net are robust to input images.

\begin{table}
\caption{Comparison of aligned/non-aligned face image AU detection}
\label{tab5}
\begin{center}

\begin{tabular}{|c|c|c|c|c|c|}
\hline
\multicolumn{3}{|c|}{With aligned faces}&\multicolumn{3}{|c|}{With non-aligned faces}\\
\hline
F1&ACC&Time(ms)&F1&ACC&Time (ms)\\
\hline
43.6\%&69.8\%&2500& 44.8\%&70.1\%&30\\
\hline
\end{tabular}
\end{center}
\end{table}

(2) DISFA dataset result

\begin{table}
\caption{F1 score on DISFA dataset}
\label{tab6}
\begin{center}
\begin{tabular}{|c|c|c|c|c|c|c|}
\hline
AU&LSVM&APL\cite{p22}&DRML\cite{p22}&FVGG&E-N&EAC\\
\hline
1&10.8&11.4&17.3&32.5&37.2&41.5\\
2&10.0&12.0&17.7&24.3&6.1&26.4\\
4&21.8&30.1&37.4&61.0&47.4&66.4\\
6&15.7&12.4&29.0&34.2&52.5&50.7\\
9&11.5&10.1&10.7&1.67&13.4&80.5\\
12&70.4&65.9&37.7&72.1&71.1&89.3\\
25&12.0&21.4&38.5&87.3&84.2&88.9\\
26&22.1&26.9&20.1&7.1&43.5&15.6\\
Avg&21.8&23.8&26.7&40.2&44.4&48.5\\
\hline
\end{tabular}
\end{center}
\end{table}

\begin{table}
\caption{Accuracy on DISFA dataset}
\label{tab7}
\begin{center}
\begin{tabular}{|c|c|c|c|c|c|c|}
\hline
AU&LSVM&APL\cite{p22}&DRML\cite{p22}&FVGG&E-N&EAC\\
\hline
1&21.6&32.7&53.3&82.7&75.1&85.6\\
2&15.8&27.8&53.2&83.6&82.5&84.9\\
4&17.2&37.9&60.0&74.1&74.5&79.1\\
6&8.7&13.6&54.9&64.2&77.4&69.1\\
9&15.0&64.4&51.5&87.1&84.0&88.1\\
12&93.8&94.2&54.6&67.8&70.1&90.0\\
25&3.4&50.4&45.6&78.6&73.8&80.5\\
26&20.1&47.1&45.3&61.7&68.6&64.8\\
Avg&27.5&46.0&52.3&74.9&75.7&80.6\\

\hline
\end{tabular}
\end{center}
\end{table}

We follow Zhao et al \cite{p22}'s setting to evaluate our approach on DISFA dataset. Since DISFA dataset is smaller in diversity and has less positive samples of AUs than BP4D, we do not directly train the model on DISFA. Instead, we use the trained model from BP4D. Since there are 3 DISFA AUs that do not exist in BP4D, we use our pretrained fine-tuned FVGG, {\color {black} E-Net and EAC Net to extract features from the DISFA images. Afterward, we use linear regression to transform our 1x2048 features to 1x8 AU prediction labels.} 27 subjects are split into 3 folds to make sure the predictions are independent.
\begin{table}
\caption{AU occurrence rates in DISFA dataset}
\label{tab8}
\begin{center}
\begin{tabular}{|c|c|c|c|c|c|c|c|}
\hline
AU1&AU2&AU4&AU6&AU9&AU12&AU25&AU26\\
\hline
4.9\%&4.3\%&15.2\%&7.8\%&4.1\%&12.88\%&27.7\%&8.8\%\\
\hline
\end{tabular}
\end{center}
\end{table}
The AU detection accuracy and F1 score are shown in Tables \ref{tab6} and \ref{tab7}. Compared to the state of the art approaches, we see more significant improvement than that with BP4D. The F1 score of the EAC Net increases by 4.1\% over the E-Net, and the average accuracy increases by 4.9\%. More importantly, the improvement yielded by C-Net is more significant than by the E-Net. This may be due to the following reasons:

(1) Training dataset balancing. The DISFA dataset is more imbalanced than BP4D. If we directly use all the raw data, the AU occurrence rates are shown in Table \ref{tab8}. Our preprocessing in balancing the data can improve the AU detection results.

(2) Our approach is more robust in dealing with wild images. The DISFA subjects have a small angle to the frontal view, so normalization is required in most approaches, while our approach merely needs to know the approximate landmarks positions on the faces. This make our approach much more robust dealing with faces not in frontal view.  

\section{Conclusions}

In this paper, we propose a region of interest based approach for AU detection. We design the enhancing net (E-Net) to force the neural network to pay more attention to AU interest regions on face images. We also propose the cropping net (C-Net) to ensure that the network learns features in ``aligned" facial areas. This makes our EAC Net -- the integration of the E-Net and C-Net -- more robust to facial shift and different orientations. 

We evaluate our proposed E-Net and EAC Net. The AU detection results shows that our approach can achieve better performance on common AU datasets. We also show the robustness of our approach to additional facial alignment preprocess. With deep pretrained models and a ``smarter" way to focus on interest regions, the proposed approach shows its power in AU detection on multiple datasets. Our approach also shows the potential to deal with ``wild" image AU detection in real time, which is our ongoing work.
In the future, we will try to find more responsive areas for the enhancing and cropping nets, as currently we manually locate the positions. In addition, we will explore integrating more temporal information into the EAC net framework to deal with the wild video AU detection problem.





\begin{thebibliography}{99}

\bibitem{p1}
Ekman, Paul, and Erika L. Rosenberg. ``What the face reveals: Basic and applied studies of spontaneous expression using the Facial Action Coding System (FACS)". {\it Oxford University Press}, USA, 1997.

\bibitem{p2}
Girshick, Ross. ``Fast r-cnn." {\it In Proceedings of the IEEE International Conference on Computer Vision}, pp. 1440-1448. 2015.

\bibitem{p3}
Li, Wei, Farnaz Abtahi, and Zhigang Zhu. ``A Deep Feature based Multi-kernel Learning Approach for Video Emotion Recognition." {\it In Proceedings of the 2015 ACM on International Conference on Multimodal Interaction}, pp. 483-490. ACM, 2015.
\bibitem{p4}
Yang, Zichao, Xiaodong He, Jianfeng Gao, Li Deng, and Alex Smola. ``Stacked attention networks for image question answering." {\it arXiv preprint} arXiv:1511.02274 (2015).
\bibitem{p5}
Zhao, Rui, Wanli Ouyang, Hongsheng Li, and Xiaogang Wang. ``Saliency detection by multi-context deep learning." {\it In Proceedings of the IEEE Conference on Computer Vision and Pattern Recognition}, pp. 1265-1274. 2015.
\bibitem{p6}
He, Kaiming, Xiangyu Zhang, Shaoqing Ren, and Jian Sun. ``Spatial pyramid pooling in deep convolutional networks for visual recognition." {\it In European Conference on Computer Vision}, pp. 346-361. Springer International Publishing, 2014.
\bibitem{p7}
Ren, Shaoqing, Kaiming He, Ross Girshick, and Jian Sun. ``Faster R-CNN: Towards real-time object detection with region proposal networks." {\it In Advances in neural information processing systems}, pp. 91-99. 2015.
\bibitem{p8}
Simonyan, Karen, and Andrew Zisserman. ``Very deep convolutional networks for large-scale image recognition." {\it arXiv preprint} arXiv:1409.1556(2014).
\bibitem{p9}
Valstar, Michel, and Maja Pantic. ``Fully automatic facial action unit detection and temporal analysis." {\it In 2006 Conference on Computer Vision and Pattern Recognition Workshop (CVPRW'06)}, pp. 149-149. IEEE, 2006.
\bibitem{p10}
Eleftheriadis, Stefanos, Ognjen Rudovic, and Maja Pantic. ``Multi-conditional Latent Variable Model for Joint Facial Action Unit Detection." {\it In Proceedings of the IEEE International Conference on Computer Vision}, pp. 3792-3800. 2015.
\bibitem{p11}
S. Koelstra, M. Pantic, and I. Y. Patras. A dynamic texturebased approach to recognition of facial actions and their temporal models. {\it IEEE Transactions onPattern Analysis and Machine Intelligence}, 32(11):1940-1954, 2010.
\bibitem{p12}
Fabian Benitez-Quiroz, C., Ramprakash Srinivasan, and Aleix M. Martinez. ``EmotioNet: An Accurate, Real-Time Algorithm for the Automatic Annotation of a Million Facial Expressions in the Wild." {\it Proceedings of the IEEE Conference on Computer Vision and Pattern Recognition}. 2016.
\bibitem{p13}
Zhao, Kaili, Wen-Sheng Chu, Fernando De la Torre, Jeffrey F. Cohn, and Honggang Zhang. ``Joint patch and multi-label learning for facial action unit detection." {\it In Proceedings of the IEEE Conference on Computer Vision and Pattern Recognition}, pp. 2207-2216. 2015.
\bibitem{p14}
Wang, Ziheng, Yongqiang Li, Shangfei Wang, and Qiang Ji. ``Capturing global semantic relationships for facial action unit recognition." {\it In Proceedings of the IEEE International Conference on Computer Vision}, pp. 3304-3311. 2013.

\bibitem{p16}
Chu, Wen-Sheng, Fernando De la Torre, and Jeffery F. Cohn. ``Selective transfer machine for personalized facial action unit detection." {\it In Proceedings of the IEEE Conference on Computer Vision and Pattern Recognition}, pp. 3515-3522. 2013.
\bibitem{p17}
Ding, Xiaoyu, Wen-Sheng Chu, Fernando De la Torre, Jeffery F. Cohn, and Qiao Wang. ``Facial action unit event detection by cascade of tasks." {\it In Proceedings of the IEEE International Conference on Computer Vision}, pp. 2400-2407. 2013.
\bibitem{p18}
Zeng, Jiabei, Wen-Sheng Chu, Fernando De la Torre, Jeffrey F. Cohn, and Zhang Xiong. ``Confidence preserving machine for facial action unit detection." {\it In Proceedings of the IEEE International Conference on Computer Vision}, pp. 3622-3630. 2015.
\bibitem{p19}
Wu, Yue, and Qiang Ji. ``Constrained joint cascade regression framework for simultaneous facial action unit recognition and facial landmark detection." {\it In Computer Vision and Pattern Recognition (CVPR)}, 2016 IEEE Conference on. 2016.
\bibitem{p20}
Girard, Jeffrey M., Jeffrey F. Cohn, L�szl� A. Jeni, Simon Lucey, and Fernando De la Torre. ``How much training data for facial action unit detection?" {\it In Automatic Face and Gesture Recognition (FG), 2015 11th IEEE International Conference and Workshops on}, vol. 1, pp. 1-8. IEEE, 2015.
\bibitem{p21}
Gehrig, Tobias, Ziad Al-Halah, Hazam Kemal Ekenel, and Rainer Stiefelhagen. ``Action unit intensity estimation using hierarchical partial least squares." {\it In Automatic Face and Gesture Recognition (FG), 2015 11th IEEE International Conference and Workshops on}, vol. 1, pp. 1-6. IEEE, 2015.
\bibitem{p22}
Zhao, Kaili, Wen-Sheng Chu, and Honggang Zhang. ``Deep Region and Multi-Label Learning for Facial Action Unit Detection." {\it In Proceedings of the IEEE Conference on Computer Vision and Pattern Recognition}, pp. 3391-3399. 2016.
\bibitem{p23}
Song, Yale, Daniel McDuff, Deepak Vasisht, and Ashish Kapoor. ``Exploiting sparsity and co-occurrence structure for action unit recognition." {\it In Automatic Face and Gesture Recognition (FG), 2015 11th IEEE International Conference and Workshops on}, vol. 1, pp. 1-8. IEEE, 2015.
\bibitem{p24}
Liu, Mengyi, Shaoxin Li, Shiguang Shan, and Xilin Chen. ``Au-aware deep networks for facial expression recognition."{\it  In Automatic Face and Gesture Recognition (FG), 2013 10th IEEE International Conference and Workshops on}, pp. 1-6. IEEE, 2013.

\bibitem{p25}
Kazemi, Vahid, and Josephine Sullivan. ``One millisecond face alignment with an ensemble of regression trees."{\it  In Proceedings of the IEEE Conference on Computer Vision and Pattern Recognition}, pp. 1867-1874. 2014.

\bibitem{p26}
He, Kaiming, Xiangyu Zhang, Shaoqing Ren, and Jian Sun. ``Deep residual learning for image recognition." {\it arXiv preprint} arXiv:1512.03385 (2015).
\bibitem{p27}
Zhang, Xing, Lijun Yin, Jeffrey F. Cohn, Shaun Canavan, Michael Reale, Andy Horowitz, Peng Liu, and Jeffrey M. Girard. ``BP4D-Spontaneous: a high-resolution spontaneous 3D dynamic facial expression database." {\it Image and Vision Computing.} Vol 32, no. 10 (2014): 692-706.
\bibitem{p28}
Lucey, Patrick, Jeffrey F. Cohn, Takeo Kanade, Jason Saragih, Zara Ambadar, and Iain Matthews. ``The extended cohn-kanade dataset (ck+): A complete dataset for action unit and emotion-specified expression." {\it In 2010 IEEE Computer Society Conference on Computer Vision and Pattern Recognition-Workshops}, pp. 94-101. IEEE, 2010.
\bibitem{p29}
Mavadati, S. Mohammad, Mohammad H. Mahoor, Kevin Bartlett, Philip Trinh, and Jeffrey F. Cohn. ``Disfa: A spontaneous facial action intensity database." {\it IEEE Transactions on Affective Computing} 4, no. 2 (2013): 151-160.
\bibitem{p30}
Valstar, Michel F., Timur Almaev, Jeffrey M. Girard, Gary McKeown, Marc Mehu, Lijun Yin, Maja Pantic, and Jeffrey F. Cohn. ``Fera 2015-second facial expression recognition and analysis challenge."{\it In Automatic Face and Gesture Recognition (FG), 2015 11th IEEE International Conference and Workshops on}, vol. 6, pp. 1-8. IEEE, 2015.

\end{thebibliography}
\end{document}